# Realistic River Image Synthesis using Deep Generative Adversarial Networks

Akshat Gautam
School of Engineering and Applied Sciences
Columbia University
ag4349@columbia.edu

Muhammed Sit
Interdisciplinary Graduate Program in Informatics
University of Iowa
muhammed-sit@uiowa.edu

Ibrahim Demir
Department of Civil and Environmental Engineering
University of Iowa
ibrahim-demir@uiowa.edu

**Abstract**
In this paper, we demonstrated a practical application of realistic river image generation using deep learning. Specifically, we explored a generative adversarial network (GAN) model capable of generating high-resolution and realistic river images that can be used to support modeling and analysis in surface water estimation, river meandering, wetland loss, and other hydrological research studies. First, we have created an extensive repository of overhead river images to be used in training. Second, we incorporated the Progressive Growing GAN (PGGAN), a network architecture that iteratively trains smaller-resolution GANs to gradually build up to a very high resolution to generate high quality (i.e., 1024x1024) synthetic river imagery. With simpler GAN architectures, difficulties arose in terms of exponential increase of training time and vanishing/exploding gradient issues, which the PGGAN implementation seemed to significantly reduce. The results presented in this study show great promise in generating high-quality images and capturing the details of river structure and flow to support hydrological research, which often requires extensive imagery for model performance.

## 1. Introduction

Remote sensing and data-driven hydrological modeling (Chen & Han, 2016) make up two significant application areas in water resources management. These fields comprise the conjunction for monitoring and analysis of water across the terrain and simulation of the streamflow in rivers and streams. Studies within this conjunction typically utilize satellite imagery in parallel with remote sensing and geographic information systems (GIS) (Choi et al., 2005) to better understand river morphology, hydrodynamics, and hydrologic processes. Subsequently, extensive data collection effort is critical to effectively model river systems. Over time, rivers develop sinusoidal, curved features called meanders (Hooke, 2007), due to the sideway flow of water and sediment that slowly erodes from the riverbanks. Because of their dynamic, ever-changing nature, they play a crucial role in the rainfall-runoff processes and assessing the true scale of a river's surrounding area to define the floodplain. For example, risk assessment for buildings within a flood-prone area is done for collaborative mitigation of flood hazards (Xu et al., 2020). The assessment data can then be relayed to first responders in the area to support rescue efforts during a flood and reduce any damage to life or property (Yildirim & Demir, 2019). Because of this very reason, it is critical for researchers and decision makers to have access to large repositories of river data (Sermet et al., 2020a).

The difficulty lies in analyzing a steady-state, long-term equilibrium that a river may converge to, since meandering rivers do not seem to have such an end behavior. As such, the curve and arc length of the meanders can be parametrized by non-linear and linear partial differential equations for fluid dynamics (Camporeale et al., 2005; Brower et al., 1984). Beyond river meandering, several hydrological modeling efforts, such as river flow detection (Gleason et al., 2014; Pavelsky, 2014), edge detection (Gupta and Liew, 2007), and exhaustive, temporal river system analysis (Gupta et al., 2002; Husain & Chaudhary, 2017), require customized data surveying, specifically river imagery, in a single region.

River imagery also holds significant importance when changes in large river systems over time are analyzed. In such a study by Gupta et al., (2002), the dynamic, unique nature of Vietnam's Mekong River is analyzed by visually observing its meanders and depths across different seasons. Due to the large magnitude of the river system, more qualitative image analysis is proposed to chart growth over time and assess the potential risks or damages any nearby construction or development projects may have on the coastline and water body. Additionally, rivers are typically large and dynamic systems which require studies beyond a small set of satellite images or time snapshots since the changes and practices on landscape, or a tributary can have significant impacts on other parts of the river (Husain & Chaudhary, 2017).

When analyzing a specific river system, synthetic imagery similar to ones from the river system becomes handy to better encapsulate the variations and general trends unique to that geographical and hydrological climate. For instance, across a given river, generating custom river images would further increase the datasets available for temporal and visual analysis. Furthermore, in urban areas and port cities where rivers are often surrounded by several buildings and man-made structures, it becomes decisive to have synthetic imagery that demonstrates the potentially adverse effects of further urban development in that area for better disaster management and recovery.

Compounding upon the width and breadth calculation of rivers, the satellite imagery is used to compute the river discharge for water bodies, given any two points separated by linear distance in a river. Gleason et al. (2014) proposed an algorithm to find the exact crossflow of a river, training their models on Landsat imagery. Another study focuses on the width-to-discharge ratio of rivers, combining the functionality of ground discharge sensors with satellite imagery to develop a model that can accurately find discharge in areas where there is a lack of data, like remote tributaries around cold, barren Arctic regions near Canada or Russia (Pavelsky, 2014). Additionally, by combining ground sensing with satellite imagery, Pavelsky (2014) circumvents the issue of clouds and weather potentially obscuring the shape and key details of rivers on the landscape.

Because of the data requirements on extensive river imagery for various research tasks, it is important to have datasets that can provide generalized visual samples to represent many use cases. In regions that may be costly or technically challenging to collect new river imagery, data-driven approaches like deep learning can produce realistic synthetic river images for potentially increasing the accuracy and effectiveness of models that rely on satellite imagery.

Many environmental models can be trained using the synthesized river image datasets. For example, a non-location-specific training of river imagery would allow for a cost-effective and time-efficient method for river edge and boundary detection to compute the width and exact curvature of rivers without using manual processing and monitoring using sensors. Generative models can significantly augment and replicate an image dataset to great detail, therefore acting as a powerful data augmentation tool for such studies. By sampling and feeding a small dataset of images across multiple seasons and years to a generative model, researchers would have an effective, wide-spanning set of river imagery datasets.

This paper explores GAN-based image synthesis methods and their applicability and performance in generating high-resolution realistic river imagery when fed with a comprising dataset. The rest of the paper is structured as follows; section 2 gives a brief summary of the literature regarding GANs, image generation, and data-driven approaches in hydrological research. Then section 3, Methodology, presents the datasets used and demonstrates how GANs were employed. Later in Section 4, the results are presented. Finally, Section 5 outlines the conclusions.

## 2. Related Work
### 2.1. Deep Learning Architectures

Deep generative learning refers to a subfield of artificial intelligence in which models learn to synthesize unique samples based on a given data distribution (Ranzato et al., 2011). This falls under the category of unsupervised learning, where the task is data-driven and the model is not given any target data besides the input dataset. For example, standard supervised learning algorithms can consist of Convolutional Neural Network (CNN) image classifiers, which analyze a pre-labeled set of images to find patterns between different annotated labels. Unsupervised generative learning, on the other hand, creates its own unique understanding of the dataset. Generative Adversarial Networks (GANs) (Goodfellow et al., 2014) are the primary structure implemented for unsupervised data generation (Sit et al., 2020). GANs consist of two separate neural networks within, a generator and a discriminator. While the generator component of the GAN produces new synthetic imagery from random vectors, the discriminator component tries to distinguish synthetic imagery from real ones, forcing the generator to be better at synthesis. The dynamics between discriminator and generator lead the way to synthesize better imagery that is closer to real imagery.

The Deep Convolutional GAN (DCGAN) (Radford et al., 2016) is the first major step taken in the field of GAN-guided image synthesis. It is a combination of a standard convolutional neural network and a feedforward neural network. Instead of working with pooling and hidden layers, the DCGAN replaces these with stridden and upsampling layers for the generator, along with reverse transpose functions for the discriminator. A latent vector z (1x100) is given as input to the generator network, and then a series of convolution layers upsample the vector to a 64x64 3-channel RGB image, mapping each latent vector given to the network to an image. Because of the DCGAN's ability to adeptly understand the latent space, Radford et al. note that the latent vector z can be modified to emphasize certain regions of an image (Salimans et al., 2016); for example, if training on face samples, the latent vector can be modified to force a picture with brown or black hair from the generator. The DCGAN does a good job of not mixing and matching parts from one image into another, so images show up with a sharp clarity on the output 64x64 resolution. One limitation stands in the scalability of the data; going from generating 64x64 to 128x128 (Karras et al., 2017) or larger images requires at least a quadratic increase in the number of data samples for the model inputs, therefore requiring much higher computational power at the expense of potentially average results. Other GAN architectures, like the Progressive Growing GAN (PGGAN) (Karras et al., 2017) or StyleGAN (Karras et al., 2019), are capable of bypassing this training time-resolution tradeoff.

The CycleGAN model performs image-to-image translation on unpaired image datasets, learning to map a series of input images with some key object to an output image; for example, the researchers developed a model that can turn a real landscape photograph into a painted portrait, given two datasets: landscapes and paintings (Zhu et al., 2017; Isola et al., 2017). The primary reason for CycleGAN's applicability to this paper is its performance in high-resolution image synthesis. The model works with a large input size and yields large output image dimensions. To circumvent the scalability and data costs associated with the standard DCGAN architecture, there are networks like StyleGAN and PGGAN. The StyleGAN replaces the generator network's latent vector input (Zhang & Sabuncu, 2018) with a separately learned input parameter, which focuses on the refinable styles of each generated image, allowing for different kinds of vector inputs between layers to generate varying styles of images at high resolutions. For example, the Celeb-A dataset (Odena et al., 2016) is a collection of celebrity face images that the StyleGAN was originally trained upon; the architecture is capable of fine-tuning specific points such as facial sharpness, nose size, jawline, hairline, etc. The StyleGAN architecture, along with the PGGAN, are capable of generating high-resolution images through a "progressively growing" GAN network structure.

## 2.2. Applications of GANs for Image Generation

ArtGAN (Tan et al., 2017) combines a DCGAN architecture with a decoder-autoencoder system that gives the generator a noise vector input along with a specific label of art style, like impressionist, natural, abstract, and such. Unlike standard images like faces or cars, portraits are often open to massive amounts of interpretation that artists fill in. Using a dataset of pre-compiled artwork with different genres and labels, the authors input a randomized label corresponding to an art form and then calculate the image loss based upon how strongly the image deviates from the rest of the training set images of that particular style. Additionally, they note that the reconstruction aspect of decoding encoded images greatly improves the accuracy and efficiency of this hybrid GAN architecture, rather than having two separate, computationally expensive generator and discriminator networks in a minimax game. When downsampling to lower resolutions, the authors utilize overlapped average pooling layers (Scherer et al., 2010), which keep the generator from using completely similar RGB pixel values across a large region of the image. To avoid a grid-like structure that usually shows up in high-resolution GANs due to upsampling layers, they used nearest neighbor upsampling (Odena et al., 2016).

To circumvent the issue of data scarcity, Wang et al. propose a CycleGAN implementation capable of taking computer graphics generated license plate images (random alphanumeric values superimposed on a colored rectangle), and converting them into photo-realistic license plate images that can then be used for image classification (2017). The trained CNN classifier is then given a few real images to further improve its accuracy and can then be used in high-speed traffic cameras for real-time license plate capture and recognition. In place of a CNN discriminator classifying the generator's output as either real or fake, they incorporated PatchGANs (Isola et al., 2017), which are miniature discriminator networks that compute the quality of being realistic for different NxN resolution image patches and then compute an average probability for the entire image. Isola et al. note that the most optimal value for N is 70, with lower values producing very inaccurate results and higher values tending to overfit and generate excessive amounts of material in the image. Using techniques from the Wasserstein GAN, they propose a novel CycleGAN architecture with a better metric of measuring image and pixel-to-pixel distances to avoid gradient explosion and mode collapse (Arjovsky et al., 2017).

Saadatnejad and Alahi develop a GAN architecture to synthesize pedestrian images in various poses on the street, which can then be used for self-driving car models to better discern pedestrians that can then be tracked and identified by the car's computer vision software (2019). The primary concern is the difficulty of tracking people in the absence or abundance of bad lighting, position changing or occluding objects. For their generator, they used a pre-trained pose model and then mapped different pedestrian images to these poses, and then these images with their corresponding poses are inputted to a decoder that tries to integrate the pedestrians' images with a custom pose. This image is then analyzed by the discriminator. Instead of using a standard generator-discriminator layout, their generator is an encoder-decoder network (Pu et al., 2016) because of the extra 'pose' label that is required for all of their images. In the future, this encoder mixed with a generator could be used for conditional image generation, fusing a set of different labels (in this case, human poses) with standard images. Because of the similarity, this task has with neural style transfer, the authors use style and perceptual losses to compare training dataset images with GAN-generated pedestrian poses.

### 2.3. Data-Driven Approaches in Hydrogical Research

Beyond the concept of image generation and generative adversarial networks, machine learning and data-driven methods have a multitude of applications in hydrology and predictive weather analytics. For example, RNNs and LSTMs (long short-term memory networks) (Tang et al., 2018; Sit & Demir, 2019; Xiang & Demir 2020; Sit et al., 2021) have been used to predict the chance of flooding based on daily water discharge and precipitation levels. Natural Language Processing (NLP) and LSTM networks are used to classify flood-related tweets from unrelated ones during Hurricane Irma (Sit et al., 2019a). In other cases, feedforward deep neural networks have been used to improve the robustness of standard weather prediction methods; for example, by inputting infrared satellite imagery, microwave scans, and images from hundreds of satellites orbiting the Earth, scientists can train neural networks to predict the levels of snow over time (Le et al., 2019) in areas more prone to map distortion, like at more extreme latitudes present in the North and South poles. Rather than just using one specific variable, like scans from a singular satellite or at a specific wavelength, deep learning enables scientists to consolidate numerous variables and use supervised neural networks and clustering algorithms (K-means) (Tse, 2010) to find novel trends. Other studies have looked at incorporating machine learning as an alternative to differential equations for modeling systems like water flow and runoff. For example, scientists have used polynomial and multivariate regression to analyze the flow of a river with non-linear behavior over differing time intervals a day (See et al., 2007). This enables them to individually analyze the value and contribution of each variable they record towards the final prediction. One another research is on increasing the spatial resolution of LiDAR models using deep generative adversarial networks (Demiray et al., 2021). Data-driven analytics applications in the field of hydrology involve crowdsourced stage measurements (Sermet et al., 2020b), generating rainfall products from NEXRAD (Seo et al., 2019), intelligent systems for flooding (Sermet & Demir, 2019), optimization of river network representation data models (Sit et al., 2019b) and crowdsourced voluntary distribution of hydrologic model computations (Agliamzanov et al., 2020).

## 3. Methodology
## 3.1. Dataset

The study presented in this paper was conducted on two separate river imagery datasets acquired through Google Earth Engine. Google Earth Engine is an online, interactive repository providing a full satellite mapping of the Earth, complete with all kinds of geological and hydrological landforms (Gorelick et al., 2017). For this study, river images were captured along different points spanning major rivers in the United States.

For the preliminary study, a relatively small-in-size dataset, namely Dataset A, 1,000 images were sampled evenly between Mississippi, Missouri, and Iowa River, and were captured via a screenshot editor in a native, 1024x1024 image resolution. To provide a streamlined file formatting template, each image was saved in the format of: "[River Name]_[Year Taken]_[Image #].jpg". Different rivers were used as sources of data so that the generative model will not stick to a specific kind of satellite background (e.g., areas with high precipitation may have more lush, green imagery in the background, whereas more arid regions may have grass resembling a yellow or brown shade). To increase the size of the dataset, we applied 9 data augmentation image transforms to the dataset by employing Fast.AI's (Fast.AI, 2021) vision library using Python, thereby generating a total of 10,000 river satellite images that forms the Dataset A for this study. The image transforms in Table 1 were applied to each of the images in both datasets.

Table 1. List of image transforms used for data augmentation.

| Image Transforms |
|---|
| Random Cropping (2x) |
| Hue/Saturation |
| Dihedral (Flip + Rotate) |
| Gaussian Noise |
| Affine Transform |
| Random Rotation (3x) |

Furthermore, a dataset that is more comprehensive in size but with limited visual variability, consisting of 110,000 data points, was curated, namely dataset B. In order to do so, 11,000 images were collected from major rivers in the United States, such as the Mississippi, Missouri, Colorado, Rio Grande, and Arkansas rivers. Then, the former 9 Fast.AI transforms that are also used for the dataset A were applied, with a total of 110,000 satellite images of a relatively uniform color and visual sampling. The visual variability of the two datasets was decided tentatively as the dataset A limits the visual efficacy of generation. This will be discussed in depth throughout the rest of this section and in the Results section. These datasets will be shared with the community through a GitHub repository to support benchmark and modeling studies (Ebert-Uphoff et al., 2017).

### 3.2. Deep Convolutional GAN (DCGAN)

DCGAN consists of a generator and discriminator network; they work against each other in a minimax, zero-sum game approach, where the former learns to generate realistic images capable of fooling the latter, which tries to differentiate between real and computer-synthesized images. The generator takes a Z dimensional latent vector as input, where Z is a vector somewhere in the Z-dimensional subspace called the latent space and feeds this through a series of convolutional layers to generate an image tensor. The discriminator takes in an image tensor as the input and runs it through a series of convolutional layers and one sigmoid layer at the end, outputting a probability to discern whether the network's tensor input belongs to the original data distribution or not (Ranzato et al., 2011; Goodfellow et al., 2014).

When training the DCGAN for 50 epochs on Dataset A, we obtained highly accurate and realistic image generation results; however, the DCGAN was only able to effectively generate 64x64 colored images, which is a very small resolution size. To be better analyzed by the human eye, we propose at least a 128x128 or 256x256 image size, which can lead to more qualitative observations. Using a standard DCGAN, we generated 128x128 images and ran into an issue where the generator loss immediately zeroed out, and the discriminator loss tended towards infinity. With each iteration, the loss remained the same, and no parameters were updated. The images had a broad resemblance to satellite images, sharing a green background; however, any detail in the river shape or contour was not present.

By using other upsampling techniques like the nearest neighbor and bilinear interpolation (Rukundo & Cao, 2012), while the output resolution does increase, a great amount of visual detail is lost, and the image becomes very blurry. Since such a standardized algorithm is applied evenly to every section of the image, a grid-like pattern seems to emerge among all of the images. On a standard computer screen, a 64x64 image is about 1 square inch. From an analytical perspective, this kind of image is quite small for the human eye to qualitatively differentiate rivers, terrain, and greenery. The actual detail of the image is limited by the number of pixels present, which warrants the need for an exponentially higher resolution. Consequently, another approach was needed to generate higher resolution satellite river imagery. Thus, a Progressive Growing GAN (PGGAN) was utilized to generate high-resolution (1024x1024) realistic river satellite imagery.

### 3.3. Progressively Growing GAN (PGGAN)

In a PGGAN (Karras et al., 2017), the basic process is to train the discriminator at a very low resolution, initially starting at 4x4 and building the model up slowly and iteratively by adding layers and fine-tuning up to exponentially larger resolutions in powers of 2. The lower resolution images are prepared by performing a center crop on the input image to get to the desired input resolution. The adaptively growing nature of the networks makes it easier for them to learn different styles and components of an image; instead of having to learn how to just map a random noise latent vector to an image of massive resolution like 512x512 or 1024x1024, the networks learn gradually by starting with a simple 4x4, 8x8, 16x16, etc. At each resolution when a GAN is trained, there is a 'fadeout' block layer that helps to smoothen out the process of upscaling or downscaling between image dimensions. Another major improvement allowing PGGAN to do so well is through Wasserstein GAN Gradient Penalty (WGGAN-GP) loss (Gulrajani et al., 2017). The gradient is heavily penalized as the variations between synthesized images and the training dataset increase. This process allows the model to converge much faster throughout resolutions and still boast a better model accuracy via the Inception score (Salimans et al., 2016), generating more realistic output images.

When training a PGGAN, the data is organized so that each successively increasing resolution size is cropped from the original dataset, and the resolution must be a power of 2. We wrote a script to find the exact coordinate center of each river satellite image and then perform a crop of a 1024x1024 grid in every cardinal direction. When capturing images in the dataset, we did not find the usage of a random crop effective since there was a high probability of the script capturing a section of image that was primarily grassland and not the actual water body we aimed to generate via PGGAN. We wrote a script to find the pixel midpoint of the image and then cropped a 1024x1024 region of the image (512 up, 512 down, 512 left, 512 left). At each resolution, the GAN would be able to access sets of images present in resolution folders like 4, 8, 16, etc. As we reached the 1024x1024 resolution, the storage size significantly grew to about 25 GB.

PGGAN architecture for custom 1024x1024 satellite image generation is given in Figure 1. Each tier/block of the image represents a resolution (power of 2) that the model is trained on. The generator is given a 512x1 latent vector input, and the discriminator is trained upon real samples to then discern the efficacy of the generator-created images. WGAN-GP loss can be pictured above for training both networks. Over time both networks converge to the point where the generator develops super-realistic images that can 'trick' the discriminator into thinking a generated sample is a real image.

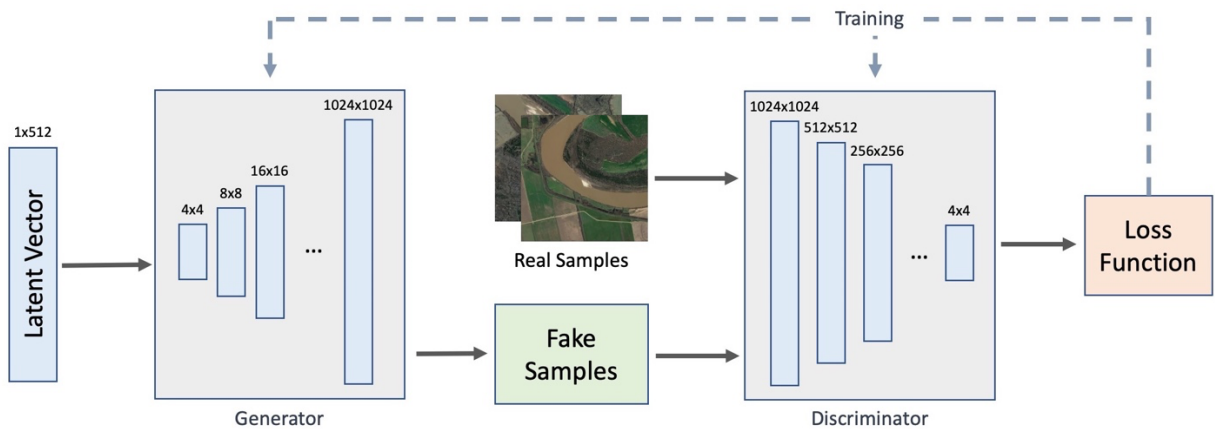

**Figure 1.** PGGAN architecture for custom 1024x1024 satellite image generation.

The PGGAN was trained on the Dataset A first, but an issue soon arose when training the PGGAN upon higher resolutions, such as 256x256 and beyond. As mentioned before, Dataset A had a very wide span of diverse kinds of river satellite images. Some were dark backgrounds full of lush greenery, while others were rivers in crowded forests, while others were in barren, sandy desert areas with little to no vegetation. As a result, it became quite difficult for the model to converge and develop images from exactly one of these types of categories each time. The PGGAN often synthesized a sample that had patches of dark vegetation and barren desert land in the same image, and we hypothesized this was due to the lack of training data; although there were 10,000 images, there was a great deal of diversity throughout the dataset. Consequently, the PGGAN was trained over Dataset B, which was curated over similar-looking rivers and is far larger in terms of the number of images.

The PGGAN model on Dataset B was trained for approximately ten days on 2 NVIDIA Titan V GPUs, each with 12 GB memory, which allowed us to converge and train faster when using a higher latent vector space (z) of 512. We trained the PGGAN architecture with a Wasserstein GAN gradient penalty loss and a mini-batch size of 16. Because of the progressive architecture, we found it important to train smaller resolution images with large iterations; 4x4 images were trained for 48,000 iterations, while 8x8 - 512x512 resolutions were trained upon 96,000 iterations. For the 1024x1024 resolution, we found it very difficult to converge at 96,000; we began seeing realistic results of convergence upon increasing it to 500,000 iterations. The alpha (parameter for the intensity of fading/switching in between upscaled resolutions) started out at 1.000 and slowly decreased down to a value of approximately 0.002 at the end of the last resolution's training. The base learning rate was 0.001, and we used an Adam optimizer (beta1 = 0, beta2 = 0.99) to train both the generator and discriminator networks. To improve the accuracy across increasing resolutions, we introduced 40,000 "fade-in" images between each upsampling (4-8, 8-16, …).

Throughout the training session, checkpoints were kept for every 48,000 iterations (for nine resolutions, a total of 18 checkpoints). After each checkpoint, a script was run to output the images generated during the specific training interval. Over time, the images became more realistic and representative of the diverse terrain in a satellite image and were able to quantitatively discern the difference between the actual river's curving path and the background landscape. To view the metrics of the generator and discriminator loss functions, we used TensorboardX, and our visual observations soon matched the progressively improving trends found in the loss curves.

## 4. Results

This section underlines how GANs generate synthetic river satellite imagery. The results include instances of generated images, their evaluation using both computer vision and statistical losses, and their overall impact. Figure 2 shows 64 images (256x256) generated by PGGAN trained on Dataset A model at iteration number 96,000. Across these 64 images, the final results show realistic river satellite imagery with little to no flaws in image quality. Specifically, this set of images was generated by the last checkpoint. An example of such an image in detail could be seen in Figure 3. The river can clearly be seen running through the center of the image, and the PGGAN has done a good job of capturing the green detail present in the surrounding landscape. Additionally, there is no sign of any failure to figure out the color space of the image; even with thousands of images of rivers with a green background, the GAN was able to understand that this specific image should not have a green background, and therefore was able to adapt to the different styles present in Dataset A.

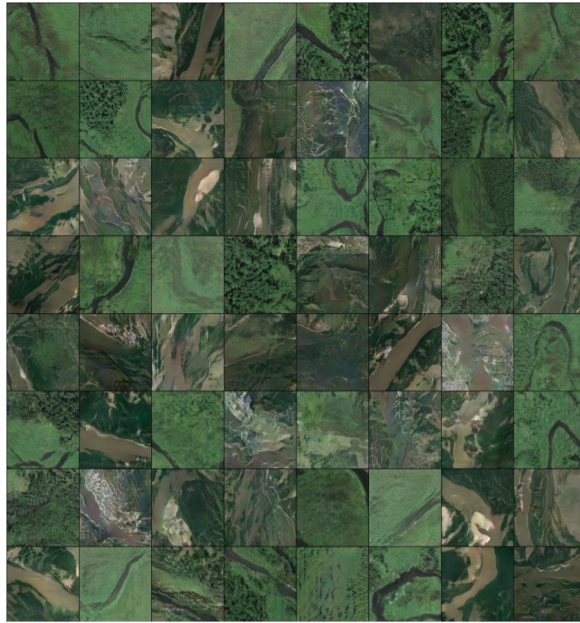

**Figure 2:** Set of 64 GAN-generated satellite river imagery at 256x256 resolution.

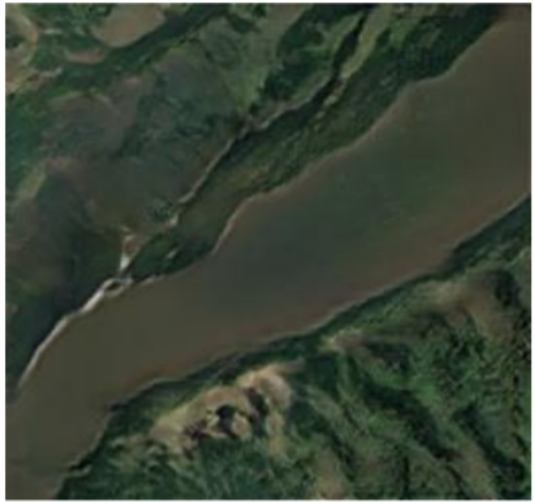

**Figure 3:** Single GAN-generated satellite river imagery at 256x256 resolution.

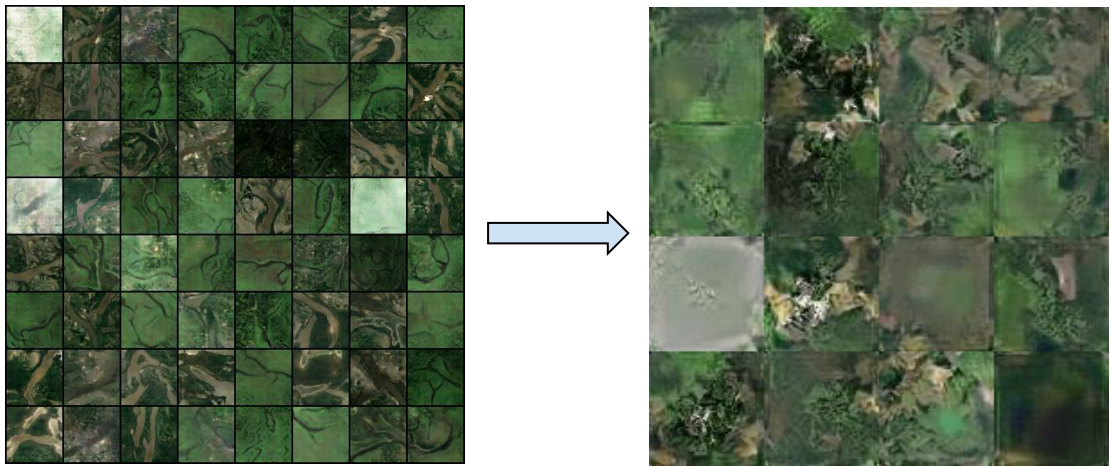

**Figure 4:** Sample of DCGAN-generated images (64x64 resolution, left), alongside the DCGAN's output for higher resolutions (right).

Figure 4 presents DCGAN's output for lower resolution imagery. The image on the left shows 64 samples of DCGAN synthesized satellite imagery, trained on a convolutional GAN for 50 epochs with a standard learning rate of 0.001, and Adam optimizers for both the generator and discriminator networks. Although the rivers seem to be highly accurate and without any imperfections, the images are of a very small size. Even with upsampling techniques like bilinear or cubic interpolation, the images will not be on par with a PGGAN-generated high-resolution sample. The image to the right shows our results when training a modified DCGAN with extra layers to produce 256x256 images. Even after 50 epochs, there are no signs of rivers forming in any of the images. This, compared with the results from our PGGAN implementation, further reinforces the benefits found in generating high-resolution satellite imagery. Figure 5 provides a set of high-resolution river satellite imagery samples generated by PGGAN. Across these 32 images, generated at the 384,000th iteration, it can be observed that the generator can recognize the various inflections in the river, as well as the sandy coast and grassy plains surrounding different kinds of water bodies.

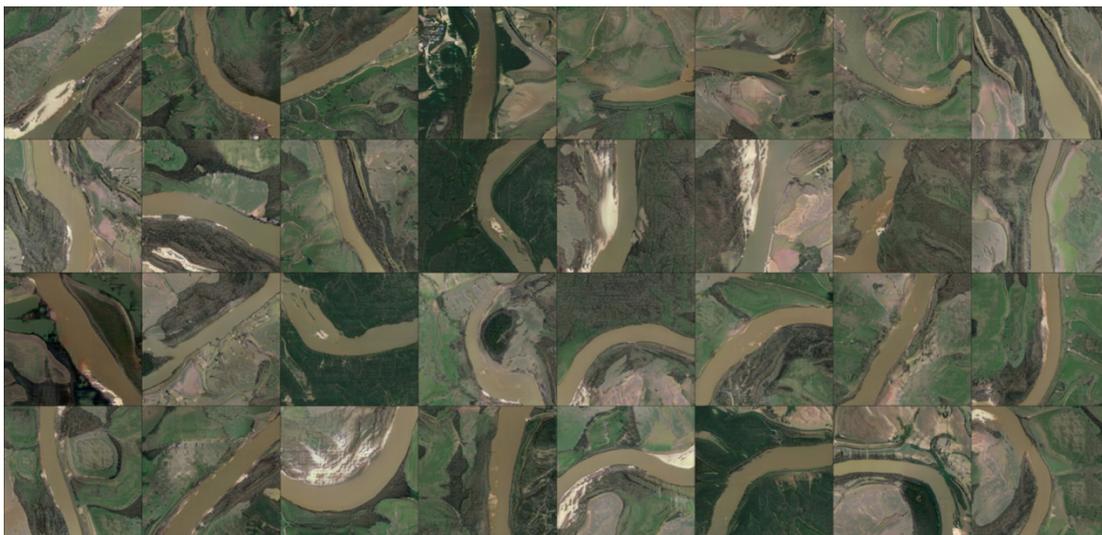

**Figure 5:** Set of 32 PGGAN-generated satellite river imagery at 1024x1024 resolution

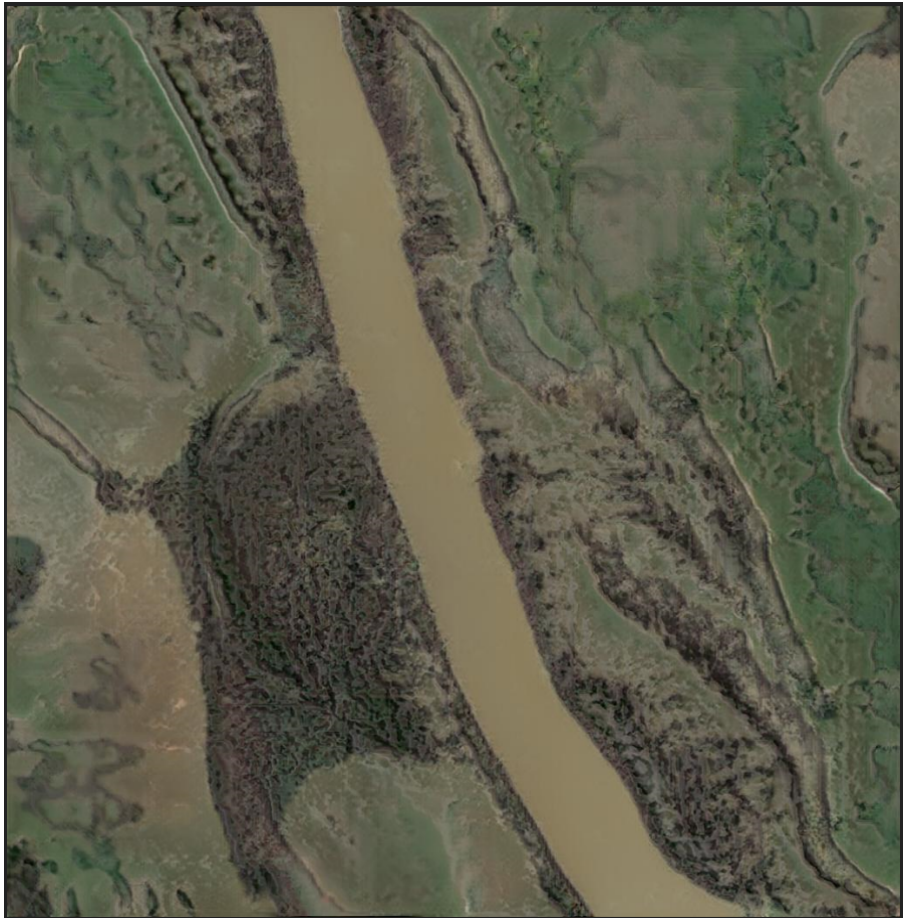
**Figure 6:** Single 1024x1024 PGGAN-generated river image sample.

Similarly, Figure 6 provides a single high-resolution PGGAN generated river imagery. The river can be seen in great detail, and the boundaries between river and land are sharper than that of the 256x256 PGGAN samples. Additionally, there is clear detail in the specificities of the land; for example, in the 1024x1024 image above, dark green areas represent overhead trees, while other sections of this image depict a mixture of green grassland and yellow plains. Conversely, in the 256x256 image, the issue of "merging" and blurring between different biomes arises, which seems to have been resolved in the new iteration of this higher resolution PGGAN.

### 4.1. Laplacian SWD
SWD, or Sliced Wasserstein Distance, is a metric that measures the overall deviation between the original training dataset and the generator-synthesized dataset (Deshpande et al., 2018). The standard Wasserstein Distance is difficult to compute on such high-dimensional input, especially in images due to the 3-color channel RGB values attached to each tensor. Each image is turned into a Laplacian pyramid, a multi-scaled data structure with layers representing each resolution that the image was generated at. To cross resolutions, there are upsampling, downsampling, and blurring functions applied to the original large image, which are then all compiled into one file. The Laplacian SWD metric converts high-dimensional inputs into one-dimensional distributions, which are randomly sampled across the image and then added up as a loss function. We computed the Laplacian SWD (Deshpande et al., 2018; Shmelkov et al., 2018) score for key resolutions across the PGGAN's training dataset (specifically resolutions from 32x32 to 256x256, Table 2, Dataset A). As seen in the table below, these normalized Wasserstein distance values tended to decrease significantly over time as the resolutions increase. This may be attributed to the PGGAN's ability to learn weights gradually across smaller resolutions and then fine-tune them when building up to image sizes greater than 64x64.

**Table 2.** The Laplacian SWD values for increasing resolutions

| Image Resolution | 32x32 | 64x64 | 128x128 | 256x256 |
|---|---|---|---|---|
| Laplacian SWD Score | 0.01337 | 0.0039 | 0.00322 | 0.00529 |

Then we trained the model to generate 1024x1024 images using PGGAN with Dataset B. As a result, we calculated the Laplacian SWD metrics for the successive resolutions, and our normalized values at these exponentially higher image sizes still tended to decrease over time (Table 3). This may have occurred due to the 500,000 training iterations we used at the highest resolution, as opposed to training everything at 96,000 iterations as was done PGGAN implementation ran over Dataset A. The Laplacian SWD scores were relatively higher at lower resolutions compared PGGAN on Dataset A; however, as the resolution grew to 512 and 1024, where the network on Dataset A was unable to converge and generate realistic images, the network on Dataset B is capable of generating very realistic river imagery with much lower normalized error metrics.

**Table 3.** The Laplacian SWD scores for the network trained on Dataset B

| Image Resolution | 128x128 | 256x256 | 512x512 | 1024x1024 |
|---|---|---|---|---|
| Laplacian SWD Score | 0.01527 | 0.01119 | 0.01111 | 0.00976 |

### 4.2. Inception Score

The inception score (IS) refers to a metric for measuring the validation and accuracy of image generation GANs. Using a pre-trained CNN Inception v3 image classifier (Szegedy et al., 2016), the algorithm classifies and categorizes the generated images into various different classes based upon the diversity and image quality of the synthesized dataset. The optimal value for an inception score is generally the number of classes present within the data. Since our data was not labeled, it becomes slightly more difficult to discern what this maximum optimal value should be. Even though the Dataset B employed in this study for PGGAN implementation only comprises images from a single type of river, the Dataset A involves different types of rivers, so a neural network trained over the Dataset A is subject to the IS. Consequently, the IS is calculated for the PGGAN trained over Dataset A. Overall, three different kinds of river images were sampled: light color rivers with desert, dark color rivers with lush greenery, and rivers with moderate greenery (Figure 7). As a result, approximately three primary river imagery categories were estimated to be in the generated and original datasets.

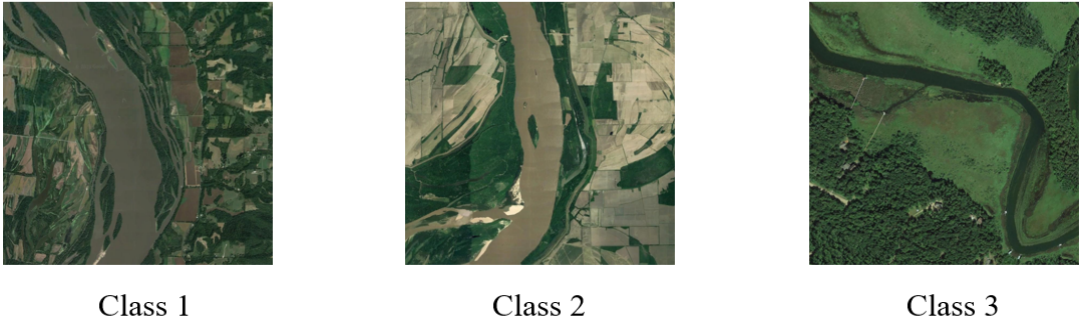

Class 1          Class 2          Class 3

**Figure 7.** Different classes of rivers with varying color distribution and greenery

The inception value for Dataset A is 2.91467. With approximately three main classes present in our data, it can be concluded that the PGGAN implementation for synthesizing custom river satellite imagery is quite efficient and accurate. Figure 7 also provides a medium to demonstrate the difference between Dataset A and B. As mentioned, Figure 7 shows three visually different river imagery, but Dataset B was designed to comprise only one type. After the experiments with Dataset A were done, Class 2 was chosen to curate the Dataset B focusing on. The primary reason for selecting Class 2 is that the river is visibly differentiable from the landscape, so the generator is capable of more easily understanding the flow and shape of a river, which is the most important part when aiming to generate realistic river imagery. Specifically, in the context of river meandering, a neural network will be more beneficial if it is able to synthesize images with sharp, detailed coastlines that can show accurate effects of erosion to a river's shape.

### 4.3. Surveying

To further analyze the performance of the model, a visual survey tool was developed to ask researchers in the field to classify a subset of 1024x1024 river satellite images as either real or fake. The surveying tool comprised a 50/50 pre-selected set of 1000 real and fake images. In the survey, participants were given a subset of 25 to 30 images and were asked to classify each of them one-by-one. To avoid adding bias to each successive question, they received the results of the survey at the end. The probability of the surveyor receiving a fake or real image was 50% at any given time, and every time a participant answered a question, a running accuracy total for that particular image was computed and stored. At the end of the survey, users received a report of the number they classified correctly and incorrectly but were not allowed to see the individual images, as this would instill bias the next time they were to take the survey. We wanted to conduct a survey such that the participants use only their own decision-making and not learn from every successive survey trial. The confusion matrix of the total of 317 classified images is provided in Table 4.

**Table 4.** Confusion matrix for surveying

|  | Predicted Real | Predicted Fake |
|---|---|---|
| **Actually Real** | 108 | 47 |
| **Actually Fake** | 49 | 113 |

The PGGAN is capable of generating many images that are difficult for participants to discern as fake. However, there was a significant number of true negatives, so people were able to correctly classify GAN-generated images as well. Therefore, there is room for improvement for river image generation; for example, the training hyperparameters could be tuned on the current dataset for better performance. Additionally, we can further modify the images in our training dataset to establish uniformity and ensure that the water body in the image is perfectly center cropped, thereby making sure that surrounding greenery and land are not prioritized over the actual river in PGGAN generation.

## 5. Conclusions

This paper presented a study to generate realistic satellite river imagery from random noise vectors using deep neural networks. While evaluating various GAN architectures such as DCGAN and PGGAN with low-resolution satellite imagery, we also explored the limits of those architectures for generating high-resolution imagery. Metrics such as Laplacian SWD were utilized for increasing resolutions of generated images. Based on those metrics and the overall visual quality of outputs of different implementations, we showed that PGGAN based implementation outperforms DCGAN. Another outcome of this study is a novel dataset for river imagery generation and augmentation of thousands of overhead river satellite images.

In the future, other GAN architectures can be utilized to improve the customizability and applicability of these river images. For example, the CycleGAN and other image-to-image translation techniques (Zhu et al., 2017; Isola et al., 2017) can be utilized to translate a set of drawings/sketches of curves into custom shaped rivers, which can be used by researchers to develop different kinds of synthetic river images for varying simulations, like computing river flow/discharge in various areas (Gleason et al., 2014; Pavelsky, 2014) and understanding the dynamics of entire river systems and tributaries to see how man-made and environmental changes can affect the landscape and ecosystem, like in the Mekong River through Vietnam (Gupta et al., 2002).

Since there is a scarcity of universal metrics for understanding a GAN's efficiency, we plan to expand our PGGAN river satellite imagery survey and distribute it to more participants (Venkatesh et al., 2020). To establish a wider variety of images, we could generate the images directly from the PGGAN on the fly, rather than sampling from a constant repository of pre-selected images. Additionally, we plan to add functionality to ask users what irregularities they see within PGGAN generated images. For example, some GAN images tend to have a checkerboard-shaped pattern in certain parts of the image, and participants could notice some common defect or blur within generated satellite images. Users could be asked to highlight the region of the image, looking for the parts most irregular or 'incorrect' to them. This could give potential insight into whether or not our specific trained model has a common trend amongst all its images, which is another potential topic of inquiry in order to further understand the 'black box' behind how PGGANs and generative models synthesize images and data samples.

The models utilized in this study can be further trained on higher resolutions to improve the efficacy of these satellite images towards different kinds of machine learning tasks. A GAN-based data augmentation platform would be very beneficial for training CNN image classifiers on satellite and terrain imagery to detect any changes in greenery, precipitation, or other GIS information. The PGGAN can be further trained on other specific datasets to support scientific studies; for instance, up to 512x512 resolution, the PGGAN can be trained on a dataset comprising rivers from all around the world to understand the structure and winding shape. For higher resolutions, a custom dataset can be slowly introduced so that the GAN architecture can quickly learn to generate images of which there may be scarce training data. Overall, the results in our study show that PGGANs are good at analyzing a set of diverse terrain datasets (satellite images with varying backgrounds, river shapes, and greenery), and can quickly learn to model such data distributions to generate realistic river satellite imagery.

## Acknowledgements
The work reported here has been possible with the support and work of many members of the Iowa Flood Center at the IIHR - Hydroscience and Engineering, University of Iowa.

## Abbreviations
**GAN: Generative Adversarial Networks**
**PGGAN: Progressively Growing GAN**